\documentclass[letterpaper]{article} 
\usepackage{aaai23}  
\usepackage{times}  
\usepackage{helvet}  
\usepackage{courier}  
\usepackage[hyphens]{url}  
\usepackage{graphicx} 
\usepackage{natbib}  
\usepackage{caption} 
\usepackage{algorithm}
\usepackage{algorithmic}
\usepackage{multirow}
\usepackage[mathscr]{euscript}
\usepackage{amsmath}
\usepackage{amssymb}
\usepackage{tabularx, booktabs}
\usepackage{subcaption}
\usepackage{animate}

\usepackage{newfloat}
\usepackage{listings}
\DeclareCaptionStyle{ruled}{labelfont=normalfont,labelsep=colon,strut=off} 
\floatstyle{ruled}
\newfloat{listing}{tb}{lst}{}
\floatname{listing}{Listing}
\title{MultiAct: Long-Term 3D Human Motion Generation\\from Multiple Action Labels}
\author {
    Taeryung Lee\equalcontrib\textsuperscript{\rm 1},
    Gyeongsik Moon\equalcontrib\textsuperscript{\rm 3},
    Kyoung Mu Lee \textsuperscript{\rm 1, \rm 2}
}
\affiliations {
    \textsuperscript{\rm 1} IPAI, \textsuperscript{\rm 2} Dept. of ECE \& ASRI, Seoul National University, Korea\\
    \textsuperscript{\rm 3} Meta Reality Labs Research\\
    trlee94@snu.ac.kr, mks0601@meta.com, kyoungmu@snu.ac.kr
}

\begin{document}

\maketitle

\begin{abstract}
We tackle the problem of generating long-term 3D human motion from multiple action labels.
Two main previous approaches, such as action- and motion-conditioned methods, have limitations to solve this problem.
The action-conditioned methods generate a sequence of motion from a single action.
Hence, it cannot generate long-term motions composed of multiple actions and transitions between actions.
Meanwhile, the motion-conditioned methods generate future motions from initial motion.
The generated future motions only depend on the past, so they are not controllable by the user's desired actions.
We present \textbf{MultiAct}, the first framework to generate long-term 3D human motion from multiple action labels.
MultiAct takes account of both action and motion conditions with a unified recurrent generation system.
It repetitively takes the previous motion and action label; then, it generates a smooth transition and the motion of the given action.
As a result, MultiAct produces realistic long-term motion controlled by the given sequence of multiple action labels.
Code is publicly available here\footnote{\url{https://github.com/TaeryungLee/MultiAct_RELEASE}}.
\end{abstract}

\section{Introduction}

Modeling and generation of realistic human motion play an essential role in computer vision and robotics, including automated avatars for AI assistant~\cite{neuhaus2019ai}, virtual reality~\cite{ahuja2021vr} and human-robot interaction~\cite{chan2021hri}.
However, despite decades of efforts to model human motions, generating controllable long-term 3D human motions remains a challenging problem.

\begin{figure}[t]
    \centering
    \includegraphics[width=\linewidth]{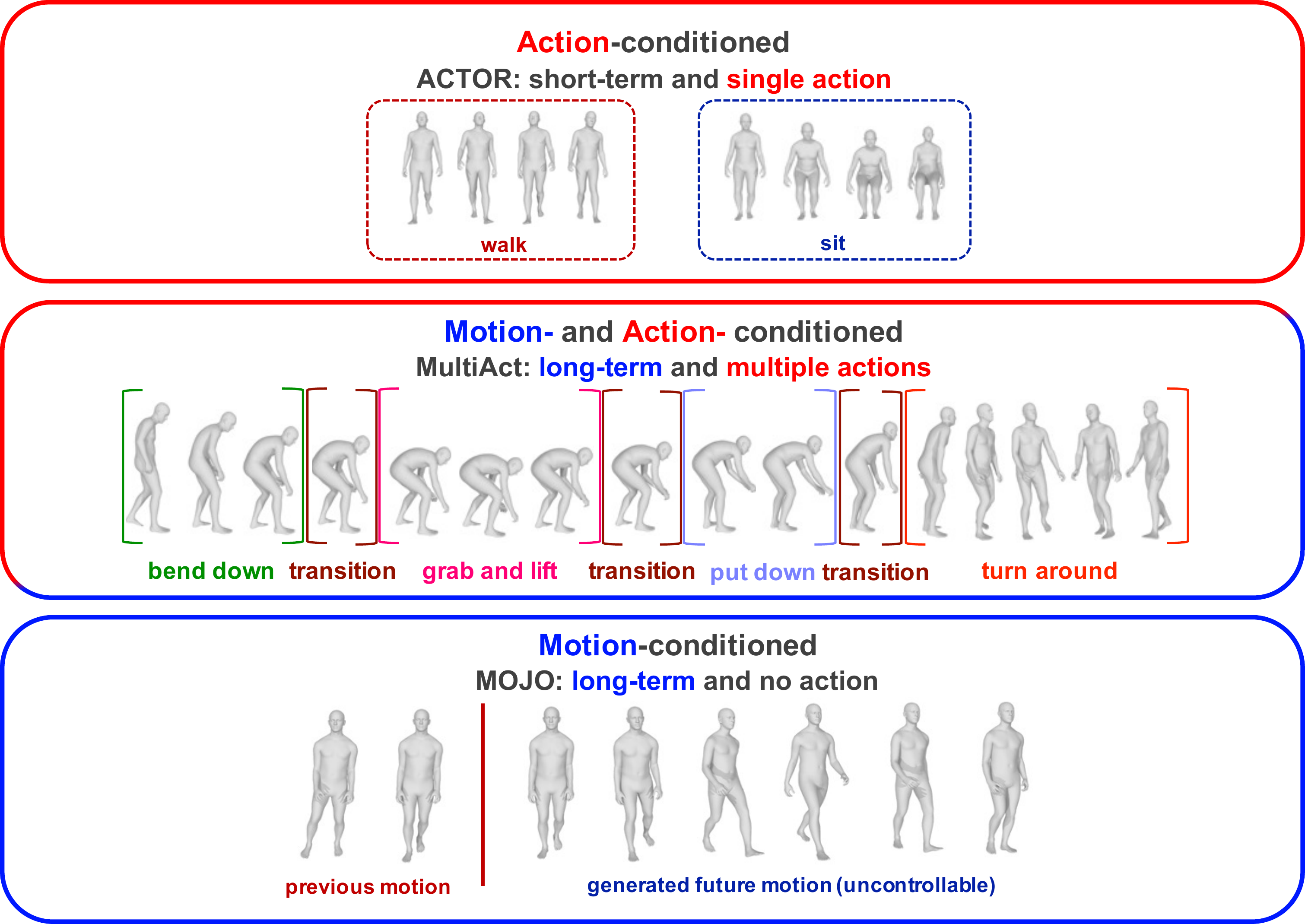}
    \caption{\fontsize{9}{10}\selectfont
    \textbf{Categorization.}
        We show long-term motions generated by our MultiAct (mid).
        Compared to ours, action-conditioned ACTOR~\cite{petrovich21actor} (top) can only generate short-term single actions.
        Motion-conditioned MOJO~\cite{zhang2021mojo} (bottom) cannot control the generated motion with desired actions.
        MultiAct handles both conditions in a single model to generate long-term motions of multiple actions.
    }
    \label{fig:fig1_modes}
\end{figure}

Fig.~\ref{fig:fig1_modes} categorizes 3D human motion generation methods by conditions used in generation.
The action-conditioned methods~\cite{Cai2018,petrovich21actor,chuan2020action2motion} generate a short-term motion from an action label, and the motion-conditioned methods ~\cite{barsoum2018hp,habibie2017recurrent,yuan2020dlow} generate future motion based on the previous motion.

However, both methods have limitations in solving our challenging target problem: \emph{``How to generate realistic long-term motion controlled by multiple actions labels?"}.
Action-conditioned methods can only produce the individual action motions, but not the realistic long-term motions composed of multiple actions and transitions between them.
Simple linear interpolation between individually generated action motions can produce multiple-action motions.
However, those interpolated transitions are unrealistic since they do not consider the adjoining motion context.
On the other hand, most motion-conditioned methods cannot control the generated motions.
Some works~\cite{wang2021scene,wang2021synthesizing,cao2020long} have tried to control the generated motion indirectly, but still, controlling the motion with a series of actions remains challenging.
Simply combining above two methods together still does not handle the target problem:
Producing an action motion with the action-conditioned method, and then generating the transition motion with the motion-conditioned method fails to generate realistic multiple-action motion since the generated transition motion is not guaranteed to be consistent with the following action motion.
This limitation motivates us to handle both conditions in a unified model.

\begin{figure*}
    \begin{center}
    \includegraphics[width=\textwidth]{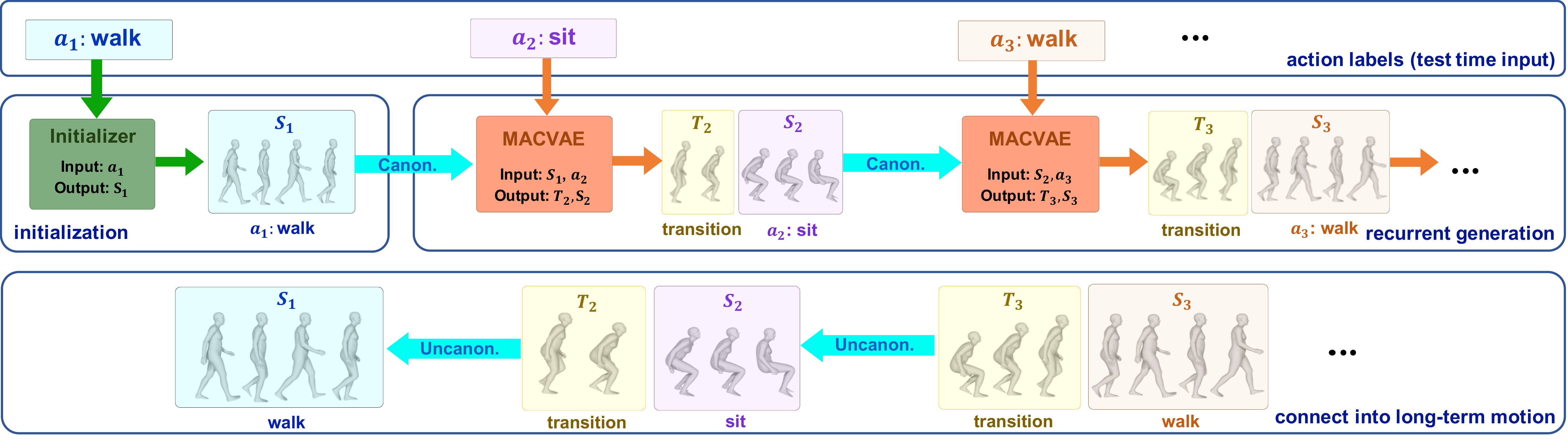}
    \caption{
        \textbf{The overview}. 
        We introduce four main steps of MultiAct to generate long-term human motion from multiple action labels.
        \textbf{[Input]} MultiAct takes a series of action labels $(a_1, a_2, ...)$.
        \textbf{[Initialization]} We use action-conditioned model to generate the initial $S_1$ from $a_1$.
        \textbf{[Recurrent generation]}
        We canonicalize the previous motion.
        Then, given the canonicalized previous motion $S_{i-1}$ and next action label $a_i$ of time step $i$, MACVAE generates the next motions ($T_{i}$, $S_{i}$).
        \textbf{[Connect into long-term motion]}
        We bring each local motion pair to the global coordinate (\textit{i.e.}, uncanonicalize) and connect it to the previous motion.
    }
    \label{fig:fig2_pipeline}
    \end{center}
\end{figure*}

We propose a novel recurrent framework, \textbf{MultiAct}, to overcome the limitations of previous approaches by handling both conditions at once.
Fig.~\ref{fig:fig2_pipeline} illustrates the overview of our framework.
MultiAct recurrently passes the previous motion and a current action label to generate the transition and the current action motion.
Fig.~\ref{fig:fig1_modes} shows an example of long-term motion from our model using the action label sequence \textit{(bend down, grab and lift, put down, turn around)}.

Two critical challenges exist in recurrently generating long-term motions from a sequence of action labels.
The first is, in each recurrent step, to generate motion that smoothly continues from the given previous motion while following the desired action.
We resolve the first challenge with a novel recurrent module MACVAE (\textbf{M}otion- and \textbf{A}ction-\textbf{C}onditioned \textbf{VAE}).
The core idea of MACVAE is to concurrently generate action motions and transitions from the joint condition of the action label and previous motion.
As a result, the generated transition is aware of the context in both adjoining motions, which is not considered in simple interpolation techniques.

The second challenge is the ground geometry losing problem during the canonicalization (\textit{i.e.}, normalization, \textit{Abbr.} canon.).
The canon. brings the previous motion into normalized form during training and testing, potentially ending in any location and facing any direction.
Such a process relieves the burden of motion-conditioned models to learn highly varying input motion space.

However, previous zero-canon.~\cite{zhang2021mojo} wipes out the global rotation that holds the geometry between the body and the ground.
Losing the information about the ground geometry leads to physical implausibility during the recurrent generation.
We adjust this problem with the face-front canon. to disentangle and retain only the relevant information from the input motion.
The experiment supports that our face-front canon. is not an ad-hoc visualization technique but an irreplaceable input normalization method used in training, single-step, and long-term testing.

To the best of our knowledge, our work is the first approach to synthesizing unseen long-term 3D human motion from multiple action labels.
We show that our MultiAct outperforms the best combination of previous SOTA methods to generate long-term motion from multiple action labels, besides handling such problem within a single model.
The experimental comparison is conducted on the quality of action motion and transition in single-step and long-term generations.
Our contributions can be summarized as follows.

\begin{itemize}
\item We propose MultiAct, a novel recurrent framework to generate long-term 3D human motion controlled by a sequence of action labels.

\item Our MACVAE concurrently generates action motions and realistic transitions aware of adjoining motion context.
Generated action motions and transitions are more realistic than previous SOTA methods.

\item Our face-front canon. assures the local coordinate system of each recurrent step shares the ground geometry.
We empirically validate the irreplaceability of the face-front canon. by qualitative and quantitative results.
\end{itemize}

\section{Related works}

\begin{figure*}[t]
\begin{center}
\includegraphics[width=.9\linewidth]{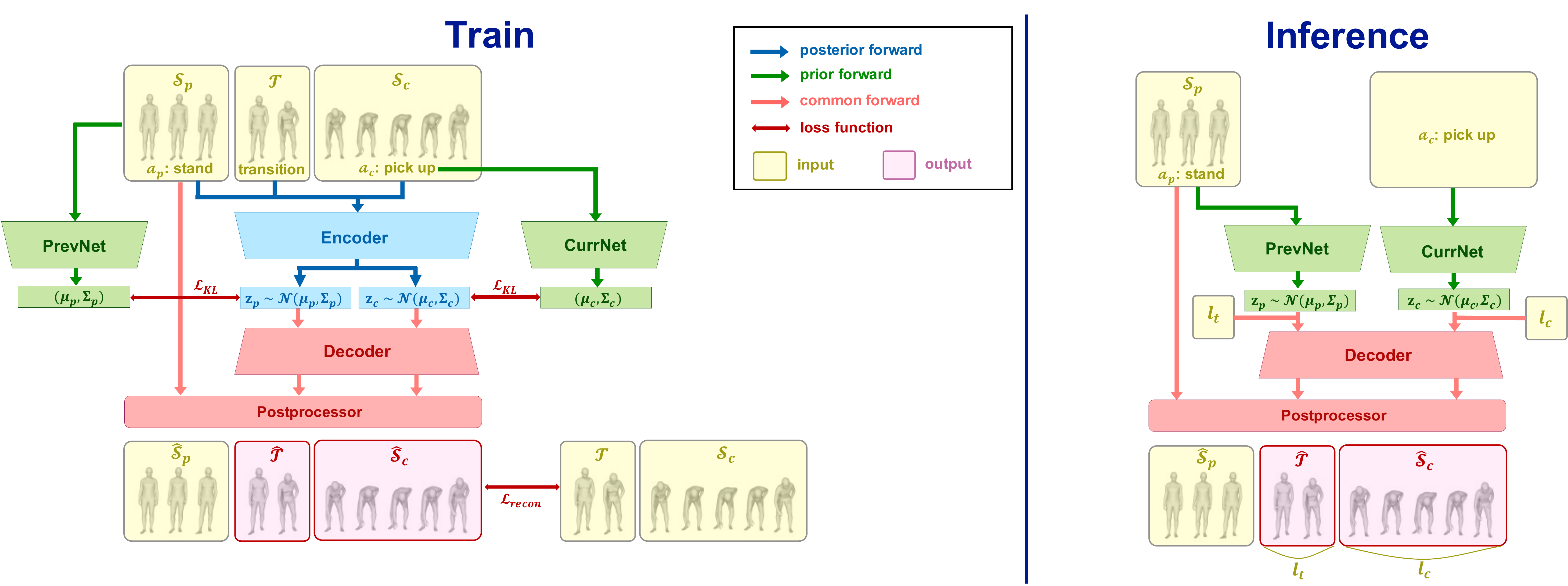}
\end{center}
   \caption{
   \textbf{Model.}
    Our model MACVAE aims to generate continuous transition and action motion $[\hat{\mathscr{T}},\ \hat{\mathscr{S}_c}]$ from previous action motion $\mathscr{S}_p$ and action label $\mathfrak{a}_c$ at inference time.
    For training, MACVAE encodes the training input $[\mathscr{S}_p,\ \mathscr{T},\ \mathscr{S}_c]$, $\mathfrak{a}_p$ and $\mathfrak{a}_c$ into posterior distribution of latent vectors $z_p$ and $z_c$ (blue).
    Through common forward of decoder and postprocessor, we reconstruct desired output $[\hat{\mathscr{T}},\ \hat{\mathscr{S}_c}]$ (pink).
    Since we do not have input motion $[\mathscr{S}_p,\ \mathscr{T},\ \mathscr{S}_c]$ at the inference time, we sample $z_p$ and $z_c$ using prior networks of PrevNet and CurrNet, respectively (green).
    We minimize divergence and reconstruction errors in training.
   }
\label{fig:fig3_model}
\end{figure*}

\noindent\textbf{Motion-conditioned human motion generation.}
Early works regressed deterministic future motions~\cite{aksan2019structured,fragkiadaki2015recurrent}.
Recently, stochastic approaches~\cite{zhang2020perpetual,chen2020dynamic} show promising results with progress in conditional generative models~\cite{sohn2015cvae}.
HP-GAN~\cite{barsoum2018hp}, DLow~\cite{yuan2020dlow} and recurrent VAE~\cite{habibie2017recurrent} predicted future motions by employing stochastic generative model.
HuMoR~\cite{rempe2021humor} and HM-VAE~\cite{li2021task} trained VAE to optimize the mesh estimation on sparse observations.
\cite{mao2021generating} proposed a model fixing the motion of the partial body and generating the motion for the remaining part.
In contrast to the previous methods that cannot control the motion~\cite{yuan2020dlow,li2021task} or indirectly controls~\cite{mao2021generating}, our method directly controls the generated motions by action labels while inheriting the spirit to generate the future from the past.

\noindent\textbf{Action-conditioned human motion generation.}
\cite{Cai2018} have presented GAN-based action-conditioned 2D human motion generation.
More recently, MUGL~\cite{maheshwari2021mugl}, Action2Motion~\cite{chuan2020action2motion} and ACTOR~\cite{petrovich21actor} proposed VAE-based action-conditioned 3D human motion generation models.
While such methods are limited to generating individual short-term motion of single action, our MultiAct can generate the long-term motion of multiple actions from joint conditions of motion and action.
PSGAN~\cite{yang2018pose} is the 2D skeleton motion generating model conditioned on both the initial pose and an action label.
In contrast to PSGAN, which generates single-action 2D motion in pixel space, MultiAct generates motion of multiple actions in 3D space.
\noindent\textbf{Motion in-betweening.}
In-betweening~\cite{harvey2020robust, zhou2020generative, duan2021ssmct} models generate the transition connecting two given motions.
Especially, the SSMCT~\cite{duan2021ssmct} is known to be the SOTA model for in-betweening.
Since our work is the first proposed method to generate long-term motion from multiple actions, we combine previous SOTA methods ACTOR~\cite{petrovich21actor} and SSMCT~\cite{duan2021ssmct} into a unified framework as a baseline to compare the quality of generated long-term multiple action motion.


\noindent\textbf{Text-conditioned human motion generation.}
A series of methods~\cite{ahn2018text2action,stoll2020text2sign,lin2018generating}, including two concurrent works~\cite{chuan2022tm2t,petrovich22temos} generate the human motion from given text.
Text-conditioned methods map the text-described continuous semantic space into the motion space using the language models.
On the other hand, the motivation of our model is to produce the smoothly connected long-term motion precisely from the sequence of discrete action labels.


\section{MACVAE}
Fig.~\ref{fig:fig3_model} shows the overview of MACVAE.
MACVAE is the recurrent unit of the overall framework MultiAct, used for generating the motion pair of (transition, action motion).
For the training, MACVAE takes the continuous motion $[\mathscr{S}_p,\ \mathscr{T},\ \mathscr{S}_c]$ in length of $L$ frames, and action labels $\mathfrak{a}_p$ and $\mathfrak{a}_c$.
We note the previous and current action motions to $\mathscr{S}_p$ and $\mathscr{S}_c$, respectively, with a transition $\mathscr{T}$ between them.
The action labels of $\mathscr{S}_p$ and $\mathscr{S}_c$ are denoted to $\mathfrak{a}_p, \mathfrak{a}_c \in A \subset \mathbb{Z^+}$, respectively.
From inputs, MACVAE is trained to reconstruct continuous transition and action motion $[\hat{\mathscr{T}},\ \hat{\mathscr{S}_c}]$, where the hat notation denotes generated output, targeting to be close to GT $[\mathscr{T},\ \mathscr{S}_c]$.
The inference stage of MACVAE takes previous action motion $\mathscr{S}_p$, an action label $\mathfrak{a}_c$, desired motion lengths $l_t$ and $l_c$ to generate future motion $[\hat{\mathscr{T}},\ \hat{\mathscr{S}_c}]$.
The generated $\hat{\mathscr{S}_c}$ is a motion that belongs to the given action label $\mathfrak{a}_c$, and generated $\hat{\mathscr{T}}$ smoothly connects in between $\mathscr{S}_p$ and $\hat{\mathscr{S}_c}$.

\subsection{Inputs and outputs}\label{subsec:3.1inputs}
\noindent\textbf{3D human motion representation.}
We represent 3D human motion of length $l$ as a sequence of 3D human pose representations $(\mathbf{p}_1, \mathbf{p}_2, ..., \mathbf{p}_l) \in \mathbb{R}^{315 \times l}$.
We note the pose representation of $i$th frame to $\mathbf{p}_i = (\mathbf{r}_i, \text{vec}(\theta_i), \mathbf{x}_i) \in \mathbb{R}^{315}$, a concatenation of global rotation $\mathbf{r}_i \in \mathbb{R}^6$, 3D joint rotations $\theta_i \in \mathbb{R}^{51 \times 6}$ and 3D translation $\mathbf{x}_i \in \mathbb{R}^3$.
Rotations $\textbf{r}_i$ and $\theta_i$ respectively represent one global rotation of the human body and the other 51 rotations of human joints, defined in the SMPL-H body model~\cite{MANO2017}, in 6D rotation~\cite{zhou2019rotation6d}.
The 3D translation $\mathbf{x}$ provides the displacement of the root joint.
Furthermore, 3D mesh vertices $V_i$ and joint coordinates $J_i$ can be obtained by forwarding the pose $\mathbf{p}_i$ to the differentiable SMPL-H layer.

\begin{figure}[t]
\centering
\begin{subfigure}{.4\textwidth}
  \centering
  \includegraphics[width=.9\linewidth]{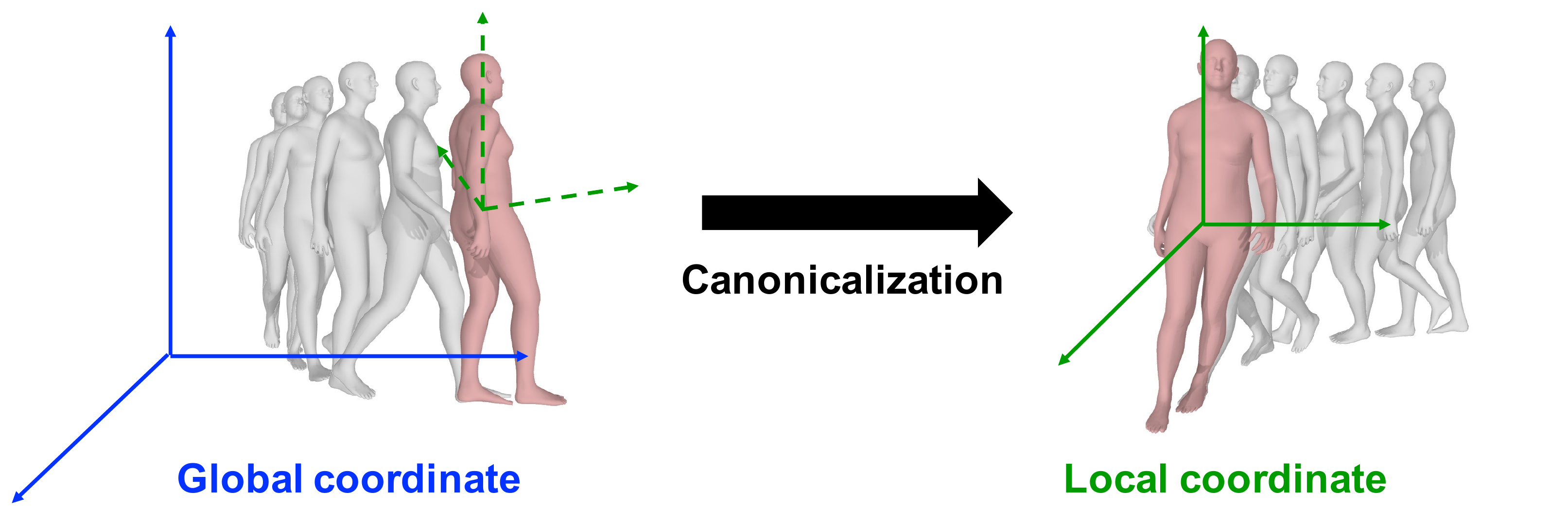}  
  \subcaption{}
  \label{fig:fig4a_normalization}
\end{subfigure}
\centering
\begin{subfigure}{.4\textwidth}
  \centering
  \includegraphics[width=.9\linewidth]{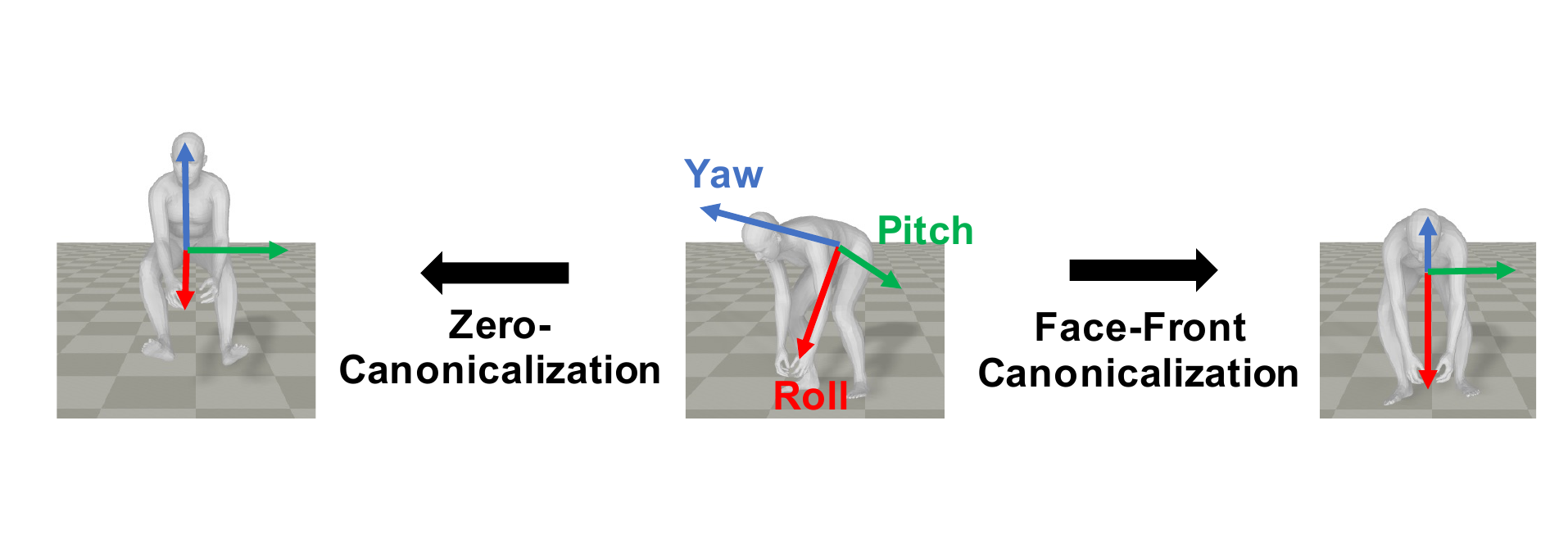}
  \subcaption{}
  \label{fig:fig4b_facefront}
\end{subfigure}
\caption{
    \textbf{Canonicalization.}
    Fig.~\ref{fig:fig4a_normalization} illustrates the face-front canon. of the previous motion.
    The last frame (red) is facing front after canon.
    Fig.~\ref{fig:fig4b_facefront} visually compares the face-front and zero-canon.
    Our face-front canon. retains the relative geometry between the body and the ground (\textit{i.e.}, how much the body is bent down) while zero-canon. does not.
}
\label{fig:fig4_normalization}
\end{figure}

\noindent\textbf{Face-front canonicalization of motions.}\label{subsubsec:3.1.2normalization}
Applying the face-front canon. to pre-canonicalized motion sequence $\{\mathbf{p}'_i = (\mathbf{r}'_i, \text{vec}(\theta'_i), \mathbf{x}'_i)\}^l_{i=1}$ with respect to anchor frame index $i_*$ formulates the canonicalized motion $\{\mathbf{p}_i\}^l_{i=1}$ in local coordinate system as following, where the local coordinate system is defined as 3D space of each recurrent step:
\begin{enumerate}
    \item Apply rotation function $f(\mathbf{r}'_i; \mathbf{r}'_{i_*}) = \mathbf{r}_i$ on global rotation $\mathbf{r}'_i$ of each frame $i$ in $\mathbf{p}'_i$.
    The function $f(*; \mathbf{r}'_{i_*}): \mathbb{R}^3 \rightarrow \mathbb{R}^3$ is a fixed rotation function uniquely determined by the global rotation $\mathbf{r}'_{i_*} = E(\alpha, \beta, \gamma)$ of anchor frame $i_*$, where $E(*, *, *)$ means the ZYX-Euler angle representation of rotation.
    The formulation of the function is $f(\mathbf{r}_{i_*}; \mathbf{r}_{i_*}) = E(\alpha, \beta, 0)$.

    \item Convert the global translation into relative translation: $\mathbf{x'_{\text{rel}, i}} = \mathbf{x'_i} - \mathbf{x'_{i_*}}$. 
    Then, we transform the relative translation $\mathbf{x'_{\text{rel}, i}}$ into translation as the local coordinate $\mathbf{x_i}$ applying the rotation $f(\mathbf{0}; \mathbf{r}'_{i_*})$. (\textit{i.e.}, align the trajectory with the viewing direction).
    
    \item Formulate the canonicalized motion $\{\mathbf{p}_i\}^l_{i=1}$ with $\{\mathbf{r}_i\}^l_{i=1}$ and $\{\mathbf{x_i}\}^l_{i=1}$, while keeping the other 51 local joint rotations $\{\theta_i\}^l_{i=1}$ the same with $\{\theta'_i\}^l_{i=1}$.

\end{enumerate}

During the training phase, we canonicalize $[\mathscr{S}_p,\ \mathscr{T},\ \mathscr{S}_c]$, so we put $l=l_p + l_t + l_c$ and anchor frame $i_*=l_p$, where $l_p, l_t, l_c$ is the respective length of $\mathscr{S}_p$, $\mathscr{T}$ and $\mathscr{S}_c$.
For the generation phase, we only canonicalize the previous motion $\mathscr{S}_p$, thus we use $l=l_p$ and anchor frame $i_*=l_p$.
In summary, face-front canonicalized motion faces front at frame $l_p$ by making the \textit{yaw} rotation in ZYX-Euler angle to zero.
Fig.~\ref{fig:fig4a_normalization} shows the visible result of canon. on previous motion.

For comparison, we use the zero-canon.~\cite{zhang2021mojo} that shares all but rotation function $f_{\text{zero}}(\mathbf{r}'_{i_*}; \mathbf{r}'_{i_*}) = E(0, 0, 0)$ instead of $f(\mathbf{r}'_{i_*}; \mathbf{r}'_{i_*}) = E(\alpha, \beta, 0)$.
Fig.~\ref{fig:fig4b_facefront} shows the visual comparison of face-front and zero-canon.
Our method normalizes the input motion, preserving the information about how the human body is bent toward the floor.
To that end, our face-front canon. guides MACVAE to generate motions with consistent ground geometry.
Effects of the canon. methods are tested and visualized in the Ablation section.

\subsection{Architecture}\label{subsec:3.2arch}
The key idea of MACVAE is to explicitly model the previous motion $\mathscr{S}_p$ and current motion $\mathscr{S}_c$ by embedding them into separate latent vectors $z_p$ and $z_c$, respectively.
To this end, we employ a CVAE architecture~\cite{sohn2015cvae}, which consists of an encoder, decoder, and additionally with two simple prior networks~\cite{wang2019story}: CurrNet and PrevNet.
Encoder encodes $[\mathscr{S}_p,\ \mathscr{T},\ \mathscr{S}_c]$, $\mathfrak{a}_p$ and $\mathfrak{a}_c$ into the posterior distribution parameters of latent vectors $z_p, z_c \in \mathbb{R}^{d}$, where $d = 512$ is an inner dimension of Transformer.
PrevNet and CurrNet estimate the prior distribution of latent vectors from only the test time inputs.
Decoder and postprocessor reconstructs the generation target $\hat{\mathscr{T}}$ and $\hat{\mathscr{S}_c}$ from the latent vectors $z_p$ and $z_c$.

\noindent\textbf{Encoder.} 
Encoder encodes all the training input $[\mathscr{S}_p,\ \mathscr{T},\ \mathscr{S}_c]$, $\mathfrak{a}_p$ and $\mathfrak{a}_c$ into parameters of Gaussian posterior distributions: $\mu^{\text{post}}_p$, $\mu^{\text{post}}_c$, $\Sigma^{\text{post}}_p$ and $\Sigma^{\text{post}}_c$.
Estimated parameters are used to sample latent vectors $z_p \sim \mathcal{N}(\mu^{\text{post}}_p, \ {\Sigma^{\text{post}}_p}^2)$ and $z_c \sim \mathcal{N}(\mu^{\text{post}}_c, \ {\Sigma^{\text{post}}_c}^2)$ by reparameterization trick~\cite{kingma2014vae}.
Transformer encoder primarily embeds the training inputs $[\mathscr{S}_p ,\ \mathscr{T} ,\ \mathscr{S}_c] \in \mathbb{R}^{L \times 315}$, $\mathfrak{a}_p$, and $\mathfrak{a}_c$ into 2D tensor of shape $\mathbb{R}^{L \times d}$.
For each frame $i \in \{1, ..., L\}$, we linearly embed the pose $\mathbf{p}_i$ into vectors of dimension $\mathbb{R}^{d/2}$.
At the same time, we learn the embedding of dimension $\mathbb{R}^{d/2}$ for each action $\mathfrak{a} \in A$ and assign the action embedding to each frame $i$, using the corresponding action label as an index.
Those vectors are concatenated into a $d$-dimensional vector and stacked into a 2D tensor of shape $\mathbb{R}^{L \times d}$.
Transformer layers encode the embedded inputs into a 2D tensor of shape $\mathbb{R}^{L \times d}$.
We pass the encoded tensor through a temporal convolution layer, take a mean along time dimension into the vector of shape $\mathbb{R}^d$ and pass it into four separate output FC layers.
Each output layers linearly estimate the parameters of Gaussian posterior $\mu^{\text{post}}_p$, $\Sigma^{\text{post}}_p$, $\mu^{\text{post}}_c$ and $\Sigma^{\text{post}}_c$, respectively.
We sample latent vectors $z_p$ and $z_c$ from estimated parameters, then pass them into the decoder.

\noindent\textbf{Decoder.}  
The decoder takes two latent vectors $z_p$, $z_c$, that hold the context of previous and following motions, respectively.
It also takes desired length of reconstructed motion $l_t$, $l_c$.
Then reconstructs $[\tilde{\mathscr{T}}, \ \tilde{\mathscr{S}_c}]$ using the transformer decoder, where $\tilde{\mathscr{T}}$ and $\tilde{\mathscr{S}_c}$ denotes reconstructed transition and current motion, respectively.
While training, we use the length of GT motions $\mathscr{T}, \mathscr{S}_c$ for $l_t, l_c$, respectively.
We provide $l_t$ and $l_c$ as input for test time generation.
The Transformer decoder takes three inputs: key, value, and query.
We build both key and value by ``expanding'' the given latent vectors $z_p$ and $z_c$ into 2D tensor of shape $\mathbb{R}^{(l_t + l_c) \times d}$.
To this end, we repetitively stack $z_p$ for $l_t$ times, then $z_c$ for $l_c$ times into 2D tensor for key and value.
The sinusoidal positional encoding is used as a query.
From inputs above, Transformer decoder outputs a sequence of $(l_t + l_c)$ vectors of dimension $\mathbb{R}^d$.
We pass the decoded vectors through a temporal convolution layer, then linearly project each $\mathbb{R}^d$ dimension vector into poses of dimension $\mathbb{R}^{315}$.
We group the first $l_t$ 3D human poses into $\tilde{\mathscr{T}} \in \mathbb{R}^{l_t \times 315}$, and following $l_c$ poses into $\tilde{\mathscr{S}_c} \in \mathbb{R}^{l_c \times 315}$.
Putting $\tilde{\mathscr{T}}$ and $\tilde{\mathscr{S}_c}$ together, decoder outputs the reconstructed motion $[\tilde{\mathscr{T}}, \ \tilde{\mathscr{S}_c}]$.

\noindent\textbf{Postprocessor.}
Postprocessor is a single 2D convolution layer of kernel size $(315 \times 5)$ to smooth the gap between previous and generated motion.
Postprocessor takes previous motion $\mathscr{S}_p$ and generated motion $[\tilde{\mathscr{T}}, \ \tilde{\mathscr{S}_c}]$.
We first concatenate them into 2D tensor $[\mathscr{S}_p,\ \tilde{\mathscr{T}}, \ \tilde{\mathscr{S}_c}] \in \mathbb{R}^{L \times 315}$, and pass through temporal convolution layer.
We discard the previous motion $\mathscr{S}_p$ part, and output the remaining $[\hat{\mathscr{T}},\ \hat{\mathscr{S}_c}]$ as a final generation.

\noindent\textbf{Prior networks: CurrNet and PrevNet.}
For inference, prior networks are used instead of an encoder: to generate the Gaussian distribution of latent vectors $z_p$ and $z_c$ from the test time inputs, $\mathscr{S}_p$ and $\mathfrak{a}_c$, respectively.
CurrNet assigns the learnable tokens $\mu^{\text{prior}}_c(\mathfrak{a}), \Sigma^{\text{prior}}_c(\mathfrak{a}) \in \mathbb{R}^d$ for each action $\mathfrak{a} \in A$, then outputs the embedded tokens $\mu^{\text{prior}}_c(\mathfrak{a}_c)$ and $\Sigma^{\text{prior}}_c(\mathfrak{a}_c)$ for given action label $\mathfrak{a}_c$.
PrevNet takes the previous action motion $\mathscr{S}_p$ as input and estimates the parameters $\mu^\text{prior}_p$ and $\Sigma^\text{prior}_p$.
Instead of using the entire $\mathscr{S}_p$, we only embed the last few frames of $\mathscr{S}_p$ into $\mu^\text{prior}_p$ and $\Sigma^\text{prior}_p$ using a single linear layer, and add the learnable token of dimension $\mathbb{R}^d$ corresponding to \textit{transition}.
This is to prevent our model from being overfitted to seen previous motions $\mathscr{S}_p$, which could result in poor generalization to unseen previous motions.

CurrNet and PrevNet are designed to estimate the latent distribution from the test time input.
However, they are divided into separate modules to impose the different roles to $z_p$ and $z_c$: $z_p$ to deliver accurate embedding of previous motions to the decoder, and $z_c$ to provide a detailed description of the action label $\mathfrak{a}_c$.
As we do not have $[\mathscr{S}_p ,\ \mathscr{T} ,\ \mathscr{S}_c]$ at test phase, we need to sample $z_p$ and $z_c$ with only test time input $\mathscr{S}_p$ and $\mathfrak{a}_c$.
Thus, we rely on prior networks to sample $z_p$ and $z_c$ at the testing phase, while the divergence between prior and posterior distributions is minimized in training. 

\subsection{Training objectives}\label{subsec:3.3obj}
\noindent\textbf{Reconstruction.}
Our first objective is to minimize L1 reconstruction losses $\mathcal{L}_V = \sum_{i=1}^{l_t+l_c}{||V_i - \hat{V}_i||_1}$ and $\mathcal{L}_P = \sum_{i=1}^{l_t+l_c}{||\mathbf{p}_i - \hat{\mathbf{p}}_i||_1}$.
We denote the GT 3D mesh vertices and pose representation of $i$'th frame to $V_i$ and $\mathbf{p}_i$, respectively.
Similarly, the predicted mesh vertices and pose of $i$'th frame are denoted to $\hat{V}_i$ and $\hat{\mathbf{p}}_i$, respectively.
In addition, we use mesh vertex acceleration loss $\mathcal{L}_{\text{acc}}$ to enhance the quality of the generated motion.
Each terms are combined into unified reconstruction loss $\mathcal{L}_{\text{recon}} = \mathcal{L}_V + \mathcal{L}_P + \lambda_\text{acc}\mathcal{L}_\text{acc}$.

\noindent\textbf{Minimizing divergence of distributions.}
The second objective is to match the prior and posterior distribution, measured by Kullback-Leibler divergence $\mathcal{L}_{\text{KL}}$.
Minimizing $\mathcal{L}_{\text{KL}}$ leads our model to reconstruct $[\hat{\mathscr{T}},\ \hat{\mathscr{S}_c}]$ with mostly from the information in $\mathscr{S}_p$ and $\mathfrak{a}_c$.
The total loss is the sum of reconstruction and divergence losses: $\mathcal{L} = \mathcal{L}_{\text{recon}} + \lambda_{\text{KL}}\mathcal{L}_{\text{KL}}$. 

\section{MultiAct}\label{sec:4multiact}
Fig.~\ref{fig:fig2_pipeline} shows overview of MultiAct.
MultiAct recurrently generates a long-term human motion $[S_1,\ T_2,\ S_2,\ ....,\ S_N]$ from series of action labels $(a_1, ..., a_N) \in A^n$, for the variable sequence length $N$.
We use the recurrent cell MACVAE, which takes previous motion $S_{t-1}$ and action $a_t$ to generate $[T_t;\ S_t]$.
Note that we denote motions in global coordinate system (\textit{i.e.}, 3D space of long-term motion) of MultiAct to $S_t, T_t$ for each timestep $t$.
The motions canonicalized into the local coordinate system that is provided into MACVAE for each step are denoted to $\mathscr{S}_p, \mathscr{T}$ and $\mathscr{S}_c$.
In each recurrent step, we give $S_{t-1}$ and $a_t$ of MultiAct into $\mathscr{S}_p$ and $\mathfrak{a}_c$ of MACVAE.
In return, generated motion $[\hat{\mathscr{T}},\ \hat{\mathscr{S}_c}]$ of MACVAE is saved as $[T_t;\ S_t]$ in MultiAct.
We show the 3-step pipeline of MultiAct below.

\begin{enumerate}
\item \textbf{Initialize.}
We first generate the initial action motion $S_1$ with separately trained ACTOR~\cite{petrovich21actor} model using the first action label $a_1$.

\item \textbf{Recurrent generation.}
We have $S_{t-1}$ from initialization or the previous recurrent step.
First, we canon. $S_{t-1}$ onto the local coordinate system and denote to $S'_{t-1}$ as described in the MACVAE section.
Second, canonicalized $S'_{t-1}$ and $a_t$ are passed into MACVAE, and generates $[T'_t;\ S'_t]$.
Generated $S'_t$ follows the action $a_t$, and $T'_t$ connects between $S'_{t-1}$ and $S'_t$.
Finally, $S'_t$ is passed to the next step as the previous uncanonicalized motion $S_t$.
We repeat this recurrence for the given sequence of actions.

\item \textbf{Connect into long-term motion.}
From the recurrent generation, we have a sequence of local motions $S_1, [T'_2;\ S'_2], ..., [T'_N;\ S'_N]$.
We connect them inductively into a global coordinate, assuming that we have connected motions $[S_1;\ T_2;\ ...;\ S_t]$ up to time step $t$.
We uncanon. $[T'_{t+1};\ S'_{t+1}]$ onto global coordinate (\textit{i.e.}, connect to the last frame of $S_{t}$) then concatenate them.
At last, we have connected long-term motion $[S_1;\ ...;\ T_{N};\ S_{N}]$, which is controlled by action labels $a_1, ..., a_N$.
\end{enumerate}

\section{Experiments}
\subsection{Datasets}\label{subsec:5.2data}
BABEL~\cite{BABEL2021} is the only dataset that consists of a long-term human motion with sequential action labels.
The set of action labels in the BABEL contains \textit{transition} as a sole action label, like other labels, such as walk and sit.
Furthermore, the action label \textit{transition} comes in between other action labels, forming an alternating sequence of action labels.
This precisely fits our goal of modeling long-term motion with alternating action motions and transitions.
We use training and validation split for training and testing of our model, respectively.
Especially for the testing, we use the \textit{unseen previous motion inputs from the test set}, which allows our experiment to demonstrate that MultiAct successfully generalizes to unseen motions.
The detail of data sampling is illustrated in supplementary materials.

\begin{table}[t]
    \centering

    \resizebox{8.5cm}{!}{
    \begin{tabular}{lcccccc}
        \toprule
        Method & $\text{FID}_\text{train}\downarrow$ & $\text{FID}_\text{test}\downarrow$ & $\text{Acc. top1}\uparrow$ & $\text{Acc. top5}\uparrow$ & $\text{Div.}\rightarrow$ & $\text{Multimod.}\rightarrow$ \\
        \midrule
        
        Real-train & $0.019^{\pm0.004}$ & $0.90^{\pm0.014}$ & $0.94^{\pm0.0014}$ & $1.00^{\pm0.001}$ & $6.87^{\pm0.124}$ & $3.29^{\pm0.058}$ \\
        \underline{Real-test} & $0.90^{\pm0.018}$ & $0.095^{\pm0.0034}$ & $0.68^{\pm0.0039}$ & $0.89^{\pm0.0035}$ & $\underline{6.75^{\pm0.045}}$ & $\underline{3.73^{\pm0.031}}$ \\
        \midrule
        
        \textbf{MACVAE (Ours)} & $0.74^{\pm0.019}$ & $0.97^{\pm0.015}$ & $\mathbf{0.64^{\pm0.009}}$ & $\mathbf{0.86^{\pm0.003}}$ & $\mathbf{6.74^{\pm0.020}}$ & $\mathbf{3.72^{\pm0.019}}$ \\
        
        \midrule
        w/o separate latent & $1.33^{\pm0.087}$ & $1.43^{\pm0.073}$ & $0.51^{\pm0.096}$ & $0.72^{\pm0.012}$ & $6.55^{\pm0.12}$ & $4.45^{\pm0.036}$ \\
        $\mathfrak{a}_c$ to PrevNet & $1.28^{\pm0.035}$ & $1.45^{\pm0.092}$ & $0.49^{\pm0.018}$ & $0.72^{\pm0.011}$ & $6.55^{\pm0.061}$ & $4.69^{\pm0.056}$ \\
        whole $\mathscr{S}_p$ to PrevNet & $0.91^{\pm0.022}$ & $1.10^{\pm0.092}$ & $0.56^{\pm0.004}$ & $0.78^{\pm0.003}$ & $6.61^{\pm0.043}$ & $4.29^{\pm0.038}$ \\
        
        \midrule
        w/o canon. & $1.21^{\pm0.036}$ & $1.12^{\pm0.026}$ & $0.52^{\pm0.0035}$ & $0.77^{\pm0.002}$ & $6.60^{\pm0.046}$ & $4.25^{\pm0.037}$ \\
        with zero-canon. & $\mathbf{0.72^{\pm0.025}}$ & $\mathbf{0.89^{\pm0.017}}$ & $\mathbf{0.64^{\pm0.018}}$ & $0.85^{\pm0.0064}$ & $6.71^{\pm0.076}$ & $4.35^{\pm0.13}$ \\

        \bottomrule
    \end{tabular}
    }
    \caption{
        \textbf{Ablation.}
        We ablate the performance of single-step MACVAE to generate action motions against alternative designs.
        The second block ablates the high-level design idea.
        Effects of canon. methods are presented in the third block.
        Symbol $\rightarrow$ means closer to the real (underlined) is better.
    }
    \label{tab:table2_ablation_action}
\end{table}

\subsection{Evaluation metrics}
We use frechet inception distance \textbf{(FID)}, action recognition accuracy \textbf{(Acc.)}, diversity \textbf{(Div.)}, and multimodality \textbf{(Multimod.)} as the measurement of the quality of the generated motions, following previous works~\cite{petrovich21actor,chuan2020action2motion}.
\textbf{FID}$_\text{train}$ and \textbf{FID}$_\text{test}$ represents distribution divergence from generated samples to training and test set, respectively.
\textbf{Acc.} measures how likely generated motions are classified to their action label by the pre-trained action recognition model.
Lower \textbf{FID} and higher \textbf{Acc.} implies the better quality of the motion.
Meanwhile, \textbf{Div.} and \textbf{Multimod.} show the variance of the generated motion across all actions and within each action, respectively.
The value closer to the real data (underlined in Tab.~\ref{tab:table2_ablation_action}) is better.
The detail of each metric is described in supplementary materials.




\subsection{Ablation study}\label{subsec:6.3ablation}
For all ablation studies, we report the performance when the previous motions are from unseen test sets.

\noindent\textbf{Separate latent embedding.}
\emph{`w/o separate latent'} in Tab.~\ref{tab:table2_ablation_action} shows that our separate latent embedding of $z_c$ and $z_p$ is highly beneficial in every evaluation metric.
Removal of separate latent embedding changes our model to unify the prior networks (\textit{i.e.}, generate $z$ from $\mathscr{S}_p$ and $\mathfrak{a}_c$), and decoder to use only $z$ to reconstruct the output motion $[\hat{\mathscr{T}},\ \hat{\mathscr{S}_c}]$.

\noindent\textbf{Inputs of PrevNet.}
\emph{`$\mathfrak{a}_c$ to PrevNet'} in Tab.~\ref{tab:table2_ablation_action} shows that the score drops when action label $\mathfrak{a}_c$ is additionally provided to PrevNet, while our original input is only the previous motion $\mathscr{S}_p$.
The PrevNet is expected to deliver the context of the previous motion to the decoder so that the decoder can smoothly connect previous and generated motions.
In this regard, the context delivered by the PrevNet should contain information about how the previous motion ends, not an action label of the previous motion, as multiple motions can correspond to the action label.
As a result, providing action label $\mathfrak{a}_c$ to PrevNet leads to inferior results.

\begin{figure}[t]
    \centering
    \includegraphics[width=0.95\linewidth]{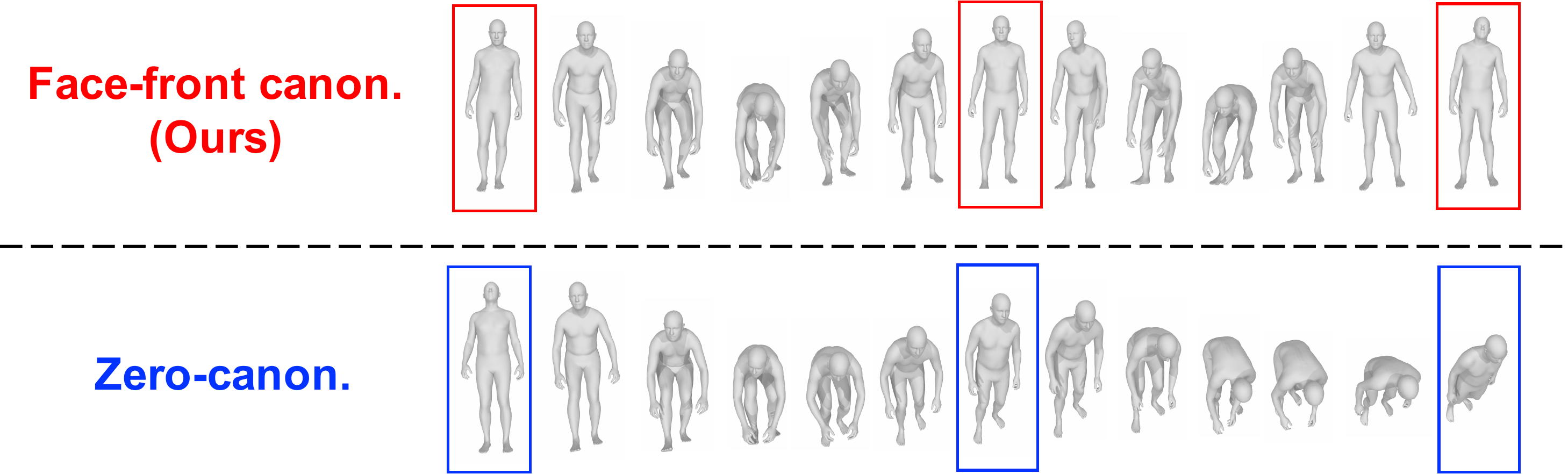}
    \caption{
    \textbf{Qualitative comparison of the canonicalization.}
    We show the long-term motion generated from \textit{(``stand", ``bend", ``stand", ``bend", ``stand")} with face-front (red) and zero-canonicalization (blue).
    }
    \label{fig:fig5_canon_result}
\end{figure}


\emph{`whole $\mathscr{S}_p$ to PrevNet'} in Tab.~\ref{tab:table2_ablation_action} shows that using whole previous motion $\mathscr{S}_p$ degrades the performance compared to ours (using last 4 frames).
Passing too many frames to PrevNet can make our system memorize unnecessary frames during the training stage.
Memorization of such unnecessary frames of the training set results in overfitting to the training set.
Our decision to use only the last four frames of the previous motion is to deliver minimal and only necessary information and to prevent the system from simply memorizing motions of the training set.

More ablations, including studies on the backbone structure and the number of frames provided to PrevNet, are provided in supplementary materials.

\noindent\textbf{Canonicalization.}
The two settings in the last block of Tab.~\ref{tab:table2_ablation_action} and Fig.~\ref{fig:fig5_canon_result} demonstrate necessity of our face-front canon.
We show that our face-front canon. plays an irreplaceable role in generating a long-term motion of multiple actions, which is more than an ad-hoc visualization method but a normalization process that has a decisive effect on both quantitative and qualitative results.
The details of canon. methods are introduced in MACVAE section and Fig.~\ref{fig:fig4a_normalization}.
Both zero-~\cite{zhang2021mojo} and face-front canon. simplify the highly varying motion space, which leads to the performance gain.

We observed that zero-canon.~\cite{zhang2021mojo} suffers from loss of the floor geometry during recurrent generation (Fig.~\ref{fig:fig5_canon_result}) since it wipes out the global roll and pitch rotation which determines the ground.
One example is the motion that ends in a bent-down position, as in Fig.~\ref{fig:fig4b_facefront}.
The zero-canon. maps the motion ``feet in the air" in the local coordinate, as illustrated in Fig.~\ref{fig:fig4b_facefront}.
As a result, generated future motion places feet back on the ground in the local coordinate.
However, in a global coordinate system (\textit{i.e.}, real-world), such motion is equivalent to leaning towards the ground. (Fig.~\ref{fig:fig5_canon_result})

As illustrated in the lower part of Fig.~\ref{fig:fig5_canon_result} and the previous paragraph, such a problem can not be handled by post-generation alignment during the visualization as the generated output is physically implausible.
Our face-front canon. is the proper normalization method that leads to the physically plausible generation result in each step (that shares the global ground geometry).
The qualitative (Fig.~\ref{fig:fig5_canon_result}) and quantitative result (last block of Tab.~\ref{tab:table2_ablation_action}) shows that our face-front canon successfully overcomes such a problem.


\begin{table}[t]
    \centering
    \setlength{\tabcolsep}{6pt}
    \resizebox{\linewidth}{!}{
    \begin{tabular}{lcccccc}
        \toprule
        Method & $\text{FID}_\text{train}\downarrow$ & $\text{FID}_\text{test}\downarrow$ & $\text{Acc. top1}\uparrow$ & $\text{Acc. top5}\uparrow$ & $\text{Div.}\rightarrow$ & $\text{Multimod.}\rightarrow$ \\
        \midrule
        Real-train & $0.019^{\pm0.004}$ & $0.90^{\pm0.014}$ & $0.94^{\pm0.0014}$ & $1.00^{\pm0.001}$ & $6.87^{\pm0.124}$ & $3.29^{\pm0.058}$ \\
        \underline{Real-test} & $0.90^{\pm0.018}$ & $0.095^{\pm0.0034}$ & $0.68^{\pm0.0039}$ & $0.89^{\pm0.0035}$ & $\underline{6.75^{\pm0.045}}$ & $\underline{3.73^{\pm0.031}}$ \\
        \midrule
        \textbf{MACVAE (Ours)} & $\mathbf{0.74^{\pm0.019}}$ & $\mathbf{0.97^{\pm0.015}}$ & $\mathbf{0.64^{\pm0.009}}$ & $\mathbf{0.86^{\pm0.003}}$ & $\mathbf{6.74^{\pm0.020}}$ & $\mathbf{3.72^{\pm0.019}}$ \\

        ACTOR & $1.48^{\pm0.043}$ & $1.53^{\pm0.032}$ & $0.50^{\pm0.0078}$ & $0.75^{\pm0.0041}$ & $6.52^{\pm0.64}$ & $4.42^{\pm0.043}$ \\
        \bottomrule
    \end{tabular}
    }
    \caption{
        \textbf{Comparison with SOTA: ACTOR.}
        We present the performance of single-step MACVAE to generate action motions from unseen previous motion and an action label.
        Previous action-conditioned SOTA ACTOR~\cite{petrovich21actor} is given as a baseline for comparison.
        Symbol $\rightarrow$ means closer to the real (underlined) is better.
    }
    \label{tab:table4_action}
\end{table}

\begin{table}[t]
    \centering

    \resizebox{5.5cm}{!}{
    \begin{tabular}{lccc}
        \toprule
        Method & $\text{FID}_\text{train}\downarrow$ & $\text{FID}_\text{test}\downarrow$ \\
        \midrule
        
        \textbf{Ours (prev. motion from testset)} & \textbf{0.87}$^{\mathbf{\pm0.052}}$ & \textbf{0.67}$^{\mathbf{\pm0.027}}$ \\
        \textbf{Ours (prev. motion from ACTOR)} & \textbf{2.20}$^{\mathbf{\pm0.67}}$ & \textbf{2.03}$^{\mathbf{\pm0.73}}$ \\
        \midrule
        
        \underline{ACTOR + SSMCT} (w. align) & \underline{6.34$^{\pm0.10}$} & \underline{5.85$^{\pm0.059}$} \\
        ACTOR + SSMCT (w. o. align) & 7.26$^{\pm0.046}$ & 7.20$^{\pm0.068}$ \\
        ACTOR + Interpolation & 13.25$^{\pm0.35}$ & 13.36$^{\pm0.29}$ \\

        \bottomrule
    \end{tabular}
    }
    \caption{\fontsize{9}{10}\selectfont
        \textbf{Comparison with SOTA: ACTOR+SSMCT.}
        We report FID score of generated transition from Ours, compared to combination of SOTA methods (ACTOR + SSMCT).
    }
    \label{tab:table3_transition}
\end{table}

\subsection{Comparison with state-of-the-art methods}
Since our MultiAct is the earliest attempt to generate long-term motion from multiple actions, we do not have directly comparable prior works.
As an alternative, we combine two SOTA models, action-conditioned ACTOR~\cite{petrovich21actor} and in-betweening SSMCT~\cite{duan2021ssmct} trained on BABEL into a unified pipeline: SSMCT connects the individually generated action motion from ACTOR.
Such SOTA-combined method is the best-known method before our work to generate multiple-action motions.
We evaluate the quality of transition and action motion in both single-step and long-term generation.

\begin{figure}[t]
    \begin{center}
    \includegraphics[width=0.77\linewidth]{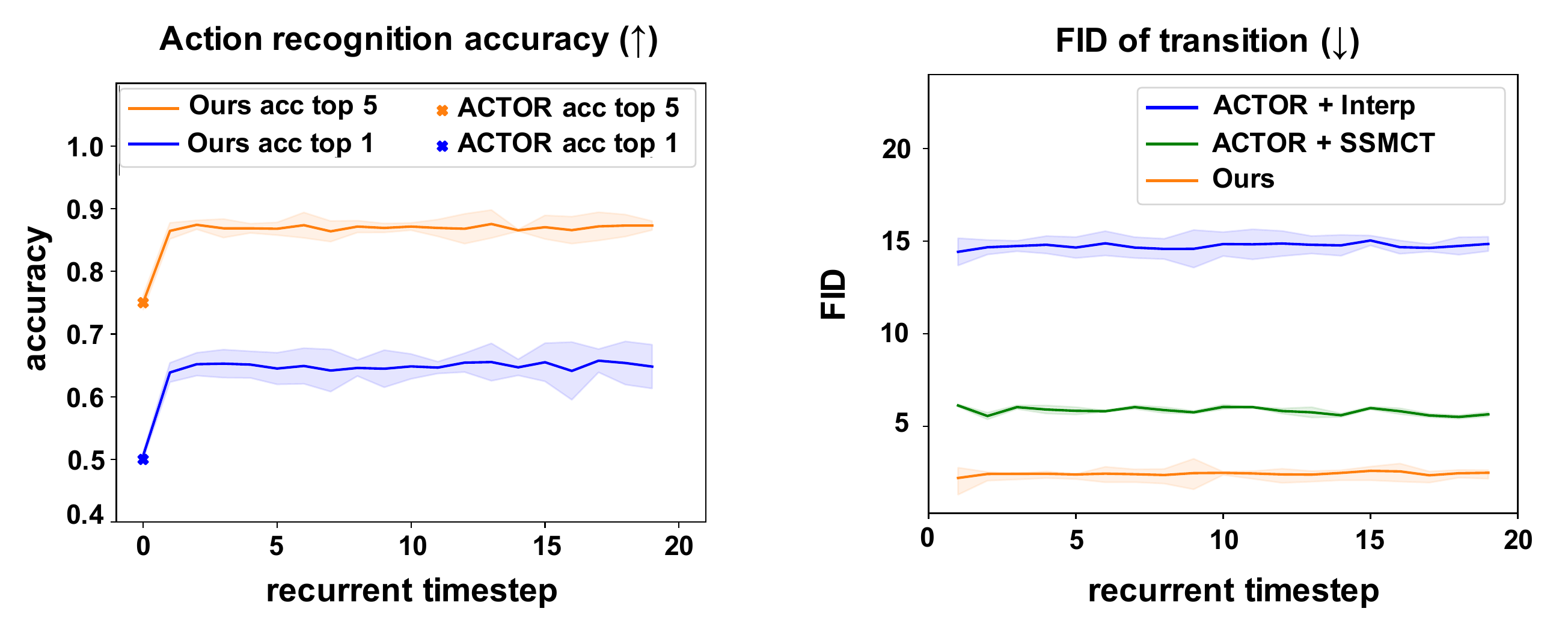}
    \end{center}
      \caption{
      \textbf{Long-term generation.}
      We report the action recognition accuracy of generated long-term motion by MultiAct (left).
      FID$_\text{test}$ of transition (right) is also reported.
      }
    \label{fig:fig6_longtermgraph}
\end{figure}

\noindent\textbf{Action motion.}
In the SOTA-combined pipeline, the generation of action motion is solely dependent on the ACTOR; thus, we compare ours to ACTOR.
In Tab.~\ref{tab:table4_action}, we have observed that single-action motions from MACVAE using the test set, \emph{which consists of unseen previous motions}, outperform the SOTA method ACTOR in all evaluation metrics.

The left plot in Fig.~\ref{fig:fig6_longtermgraph} shows that our generated long-term motion shows much higher recognition accuracy than ACTOR and maintains the quality after twenty steps of repetitive recurrence.
Note that we have sampled the input action sequence from the test set so that such a result shows that MultiAct generalizes well to the unseen permutation of action labels that are not provided during the training.
Note that MultiAct uses ACTOR to initialize $S_1$; thus, the metric score of the first step is identical to ACTOR.


\begin{figure}[t]
\begin{center}
\includegraphics[width=\linewidth]{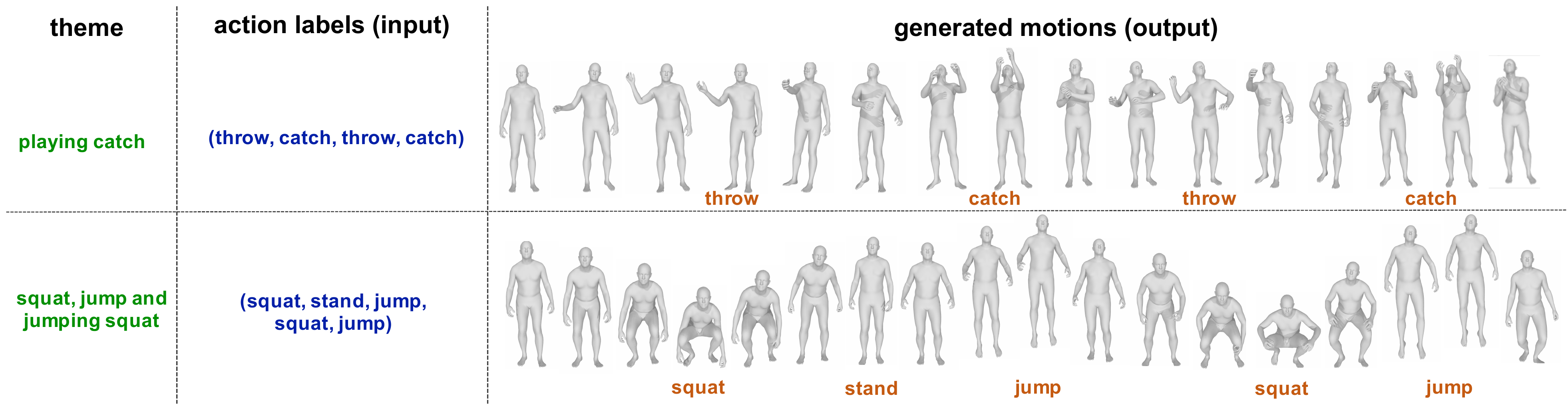}
\end{center}
    \caption{
    \textbf{Long-term motions by MultiAct.}
    We use action labels (blue) to generate the long-term motions. 
    Corresponding labels of the frame (orange) are noted below.
    Theme (green) indicates the abstracted subject of overall motion.
    }
\label{fig:fig8_long}
\end{figure}



\noindent\textbf{Transition.}
Tab.~\ref{tab:table3_transition} compares the quality of our single-step transition against SOTA-combined pipeline: SSMCT-generated transition in between two ACTOR-generated action motions.
The quality of our transition in both previous motion conditions largely outperforms both the SOTA-combined method and linear interpolation baseline.
As ACTOR generates motions independent of previous motions, the initial status of generated motions can be very different from the last one of previous motions (\textit{e.g.}, seeing the opposite direction), which can result in unnatural in-betweening of SSMCT.
The table shows that although we fix this issue by manually aligning ACTOR's action motions before performing in-betweening with SSMCT, ours still largely outperforms it.
We also report the long-term transition quality of MultiAct on the right of Fig.~\ref{fig:fig6_longtermgraph}.
Our method outperforms the other baselines throughout twenty steps of recurrence.

The performance of the SOTA-combined pipeline heavily depends on the first step, ACTOR, since the second step, SSMCT, takes the ACTOR-generated motions as an input.
The lower quality of ACTOR-generated motions (as shown in Tab.~\ref{tab:table4_action}) leads to the inferior transition quality in the second step.
The result supports our idea of simultaneously generating transition and action motion from the joint condition of motion and action.
To summarize, our MultiAct outperforms the previous SOTA on the proposed task: to generate long-term motion from multiple actions.

\subsection{Qualitative results}
Fig.~\ref{fig:fig8_long} shows our method generates highly realistic 3D human motions.
We state the abstracted theme of the generated motion (green) and action labels (blue) on the left.
Generated motions from action labels are illustrated at the right, and the corresponding action labels of the frame (orange) are stated below the motion.
Visualizations qualitatively support that our method well-generalizes to an unseen action sequence that does not come from the training or the test set, as the jumping squat example in Fig.~\ref{fig:fig8_long}.
Although the training set does not include the consecutive (squat, jump) action sequence, the generated motion performs the jumping squat just as the theme intended.
Please refer to the supplementary videos for more results.


\section{Conclusion}
We present MultiAct, the first framework to generate long-term 3D human motion of multiple actions recurrently. 
For the recurrent generation of long-term motion composed of transitions and actions, our model concurrently generates transition and action motion from the joint condition of action and motion.
As a result, our MultiAct has outperformed the previous SOTA-combined method in generating long-term motion of multiple actions.

\section*{Acknowledgements}
This work was supported in part by the IITP grants [No.2021-0-01343, Artificial Intelligence Graduate School Program (Seoul National University), No.2022-0-00156, No. 2021-0-02068, and No.2022-0-00156], and the NRF grant [No. 2021M3A9E4080782] funded by the Korea government (MSIT), and AIRS Company in Hyundai Motor Company \& Kia Corporation through HMC/KIA-SNU AI Consortium Fund.

\begin{center}
\textbf{\large Supplementary Material for\\``MultiAct: Long-Term 3D Human Motion Generation from Multiple Action Labels"}
\end{center}

\setcounter{figure}{0}
\setcounter{table}{0}
\setcounter{section}{0}

\renewcommand{\thefigure}{\Alph{figure}}
\renewcommand{\thetable}{\Alph{table}}
\renewcommand{\thesection}{\Alph{section}}

In this supplementary material, we provide more ablations and visualizations, reports of the failure cases, and the details of our experiment.






\begin{table*}[h]
    \centering

    \resizebox{16cm}{!}{
    \begin{tabular}{lcccccc}
        \toprule
        Method & $\text{FID}_\text{train}\downarrow$ & $\text{FID}_\text{test}\downarrow$ & $\text{Acc. top1}\uparrow$ & $\text{Acc. top5}\uparrow$ & $\text{Div.}\rightarrow$ & $\text{Multimod.}\rightarrow$\\
        \midrule
        Real-train & $0.019^{\pm0.004}$ & $0.90^{\pm0.014}$ & $0.94^{\pm0.0014}$ & $1.00^{\pm0.001}$ & $6.87^{\pm0.124}$ & $3.29^{\pm0.058}$ \\
        \underline{Real-test} & $0.90^{\pm0.018}$ & $0.095^{\pm0.0034}$ & $0.68^{\pm0.0039}$ & $0.89^{\pm0.0035}$ & $\underline{6.75^{\pm0.045}}$ & $\underline{3.73^{\pm0.031}}$ \\
        \midrule
        2 frames to PrevNet & $0.82^{\pm0.017}$ & $1.04^{\pm0.032}$ & $0.62^{\pm0.0059}$ & $\mathbf{0.86^{\pm0.003}}$ & $6.77^{\pm0.028}$ & $3.86^{\pm0.029}$\\
        3 frames to PrevNet & $0.80^{\pm0.034}$ & $0.98^{\pm0.037}$ & $\mathbf{0.65^{\pm0.009}}$ & $\mathbf{0.86^{\pm0.003}}$ & $6.77^{\pm0.025}$ & $3.66^{\pm0.032}$\\
        4 frames to PrevNet (Ours)& $\mathbf{0.74^{\pm0.019}}$ & $\mathbf{0.97^{\pm0.015}}$ & $0.64^{\pm0.009}$ & $\mathbf{0.86^{\pm0.003}}$ & $\mathbf{6.74^{\pm0.020}}$ & $\mathbf{3.72^{\pm0.019}}$ \\
        5 frames to PrevNet & $0.77^{\pm0.036}$ & $1.02^{\pm0.067}$ & $0.64^{\pm0.009}$ & $\mathbf{0.86^{\pm0.003}}$ & $6.71^{\pm0.01}$ & $3.66^{\pm0.035}$\\
        6 frames to PrevNet & $0.76^{\pm0.045}$ & $0.95^{\pm0.087}$ & $0.63^{\pm0.006}$ & $\mathbf{0.86^{\pm0.003}}$ & $6.71^{\pm0.021}$ & $3.69^{\pm0.045}$\\
        \bottomrule
    \end{tabular}
    }
    \caption{
        \textbf{Ablation: Number of frames given to PrevNet.}
        We ablate the performance of single-step MACVAE to generate action motions with the number of the frames passed to PrevNet.
    }
    \label{tab:table_s1}
\end{table*} 

\begin{table}[h]
    \centering

    \resizebox{8.5cm}{!}{
    \begin{tabular}{lcccccc}
        \toprule
        Method & $\text{FID}_\text{train}\downarrow$ & $\text{FID}_\text{test}\downarrow$ & $\text{Acc. top1}\uparrow$ & $\text{Acc. top5}\uparrow$ \\
        \midrule
        
        
        \textbf{Transformers (Ours)} & $\mathbf{0.74^{\pm0.019}}$ & $\mathbf{0.97^{\pm0.015}}$ & $\mathbf{0.64^{\pm0.009}}$ & $\mathbf{0.86^{\pm0.003}}$\\
        
        \midrule
        LSTM & $2.00^{\pm0.122}$ & $1.99^{\pm0.052}$ & $0.31^{\pm0.0138}$ & $0.52^{\pm0.0175}$\\
        GRU & $2.63^{\pm0.075}$ & $2.92^{\pm0.104}$ & $0.45^{\pm0.046}$ & $0.62^{\pm0.031}$\\
        MLP & $3.40^{\pm0.102}$ & $3.67^{\pm0.189}$ & $0.39^{\pm0.0068}$ & $0.56^{\pm0.0056}$\\
        
        \bottomrule
    \end{tabular}
    }
    \caption{
        \textbf{Ablation on backbone architecture.}
        We ablate the performance of single-step MACVAE to generate action motions against alternative backbone architectures.
    }
    \vspace{-4mm}
    \label{tab:table_s2}
\end{table} 

\section{More ablations}
\noindent\textbf{The number of input frames of PrevNet.}
Tab.~\ref{tab:table_s1} shows the quantitative evaluation of the number of input frames passed into PrevNet.
Among the candidate numbers of two to six, the best score was recorded when passing four frames to PrevNet.
To that end, our decision to use only the last two frames from the previous motion $\mathscr{S}_p$ in the PrevNet prevents our system from memorizing unnecessary frames during the training stage, thus beneficial to the generalization performance while given the unseen previous motions.

\noindent\textbf{Backbone architectures.}
We ablate on the backbone structure and the postprocessor in the second block of Tab.~\ref{tab:table_s2}.
Transformers have outperformed Long Short-Term Memory (LSTM), Gated Recurrent Unit (GRU), and Multi-Layer Perceptron (MLP) with similar network specifications, strengthening our choice to use Transformers to process sequential data.

\section{More visualized results}
\begin{figure*}[t]
    \begin{center}
    \includegraphics[width=0.95\linewidth]{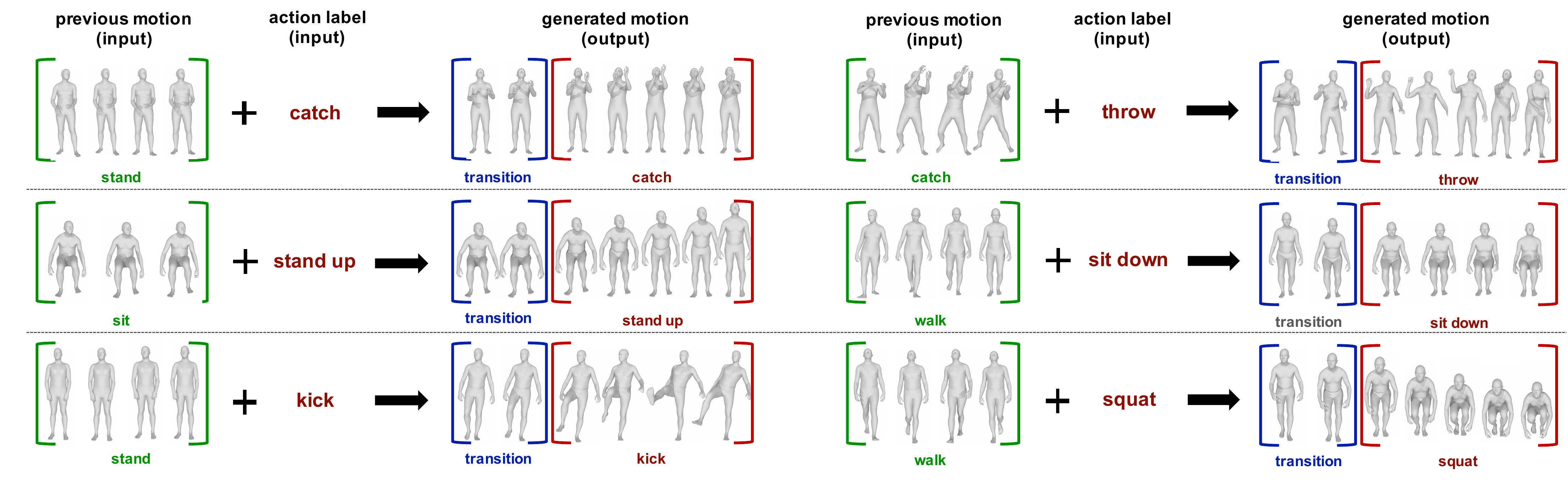}
    \vspace{-5mm}
    \end{center}
      \caption{
        \textbf{Generated single-step action motion by MACVAE.}
        We illustrate the qualitative result of generated transition (blue bracket) and action motion (red bracket) from previous motion (green bracket) and action label (red text).
        }
    \label{fig:fig_s1_single}
    \vspace{-5mm}
\end{figure*}

\begin{figure*}[t]
\begin{center}
\includegraphics[width=0.95\linewidth]{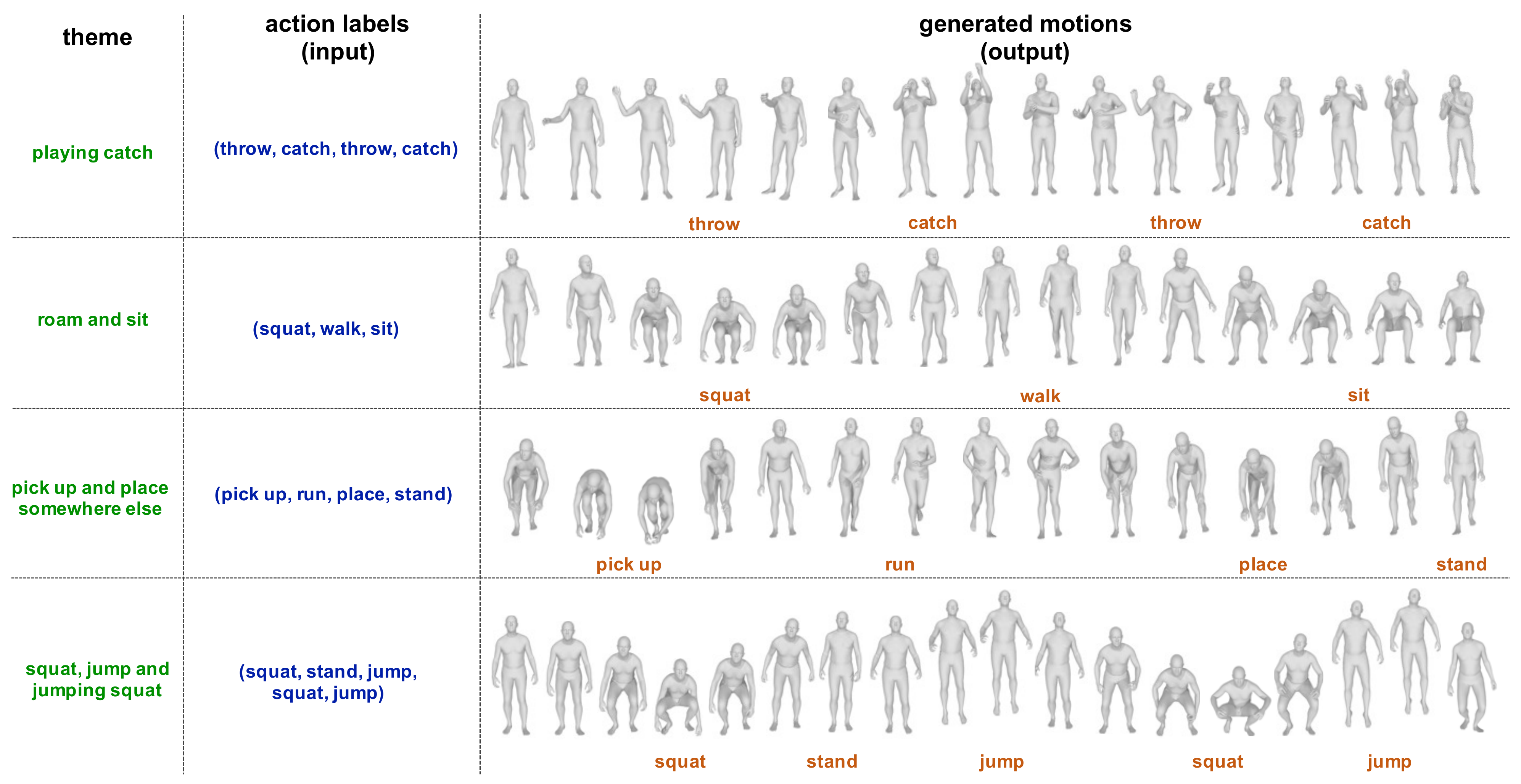}
\end{center}
    \vspace{-2mm}
    \caption{
    \textbf{Generated long-term motions by MultiAct.}
    We use action labels (blue) to generate the long-term motions. 
    Corresponding labels of the frame (orange) are noted below.
    Theme (green) indicates the abstracted subject of overall motion.
    }
\label{fig:fig_s2_long}
\vspace{-5mm}
\end{figure*}

Figs.~\ref{fig:fig_s1_single} and \ref{fig:fig_s2_long} shows our method generates highly realistic 3D human motions.
Fig.~\ref{fig:fig_s1_single} shows the single-step generation result of MACVAE.
Green, blue and red brackets respectively represent the previous motion, generated transition, and generated action motion.
We visualize generated long-term motion by MultiAct in Fig.~\ref{fig:fig_s2_long}.
We state the abstracted theme of the generated motion (green) on the left and action labels at the center (blue).
Generated motions from action labels are illustrated at the right, and the corresponding action labels of the frame (orange) are stated below the motion.


\section{Failure cases}
In this section, we report the failure cases of our MultiAct.

\subsection{Penetration} 
The first failure case is the penetration of body meshes in action labels of sit and squat.
Arms are naturally placed on legs while sitting and squatting in real-world human motion.
However, our model does not give a penetration penalty during training, so the arms of the generated motions easily penetrate through the legs while sitting and squatting.

\subsection{Locomotions} 
Second, our model sometimes fails to model locomotions such as running and jogging, especially the cyclic movement of arms and legs.
Most of the actions we use to train our model are simple one-way actions, for example, \textit{throw, kick}, and \textit{catch}.
On the other hand, running and jogging are cyclic actions that must repeat the movements of arms and legs many times.
As a result, we have observed that the generated running and jogging motion acts like running at the start but fails to repeat the running action after some period.
We suppose this limitation can be handled using a dedicated module for the cyclic motions as \textit{run}, and \textit{jog}.

\subsection{Sliding feet}
Similarly to the previous action-conditioned motion generation methods such as Action2Motion~\cite{chuan2020action2motion} and ACTOR~\cite{petrovich21actor}, MultiAct does not explicitly models the ground contact.
As a result, some of the motions generated by our model suffer from the ``sliding feet'' problem.
The post-generation ground-contact optimization can handle this problem in a similar way that is proposed by the HuMoR~\cite{rempe2021humor}.

\section{Experiment details}

\subsection{Implementation details}
We used PyTorch~\cite{paszke2017automatic} Transformer implementation for our encoder and decoder, with 512 inner dimensions, 8 layers, and 4 multi-head attentions.
We trained our model for 1300 epochs using AdamW~\cite{adamw} optimizer with a mini-batch size of 16, a learning rate of 5$\times10^{-5}$, acceleration loss weight of $\lambda_{\text{acc}} = 1$ and a KL-loss weight of $\lambda_{\text{KL}} = 10^{-5}$.
For the training and testing, we selected 53 action labels from BABEL~\cite{BABEL2021}, which appears more than 20 times in the training split.
We formed the training set by sampling 20 motions per action label.
A detailed dataset sampling method is described below in the Dataset sampling details section.

\subsection{Dataset sampling details}
\noindent\textbf{Training data for MACVAE.}
For training, we sample continuous motion $[\mathscr{S}_p,\ \mathscr{T},\ \mathscr{S}_c]$ formed with two consecutive action motions $\mathscr{S}_p$ and $\mathscr{S}_c$, and a transition $\mathscr{T}$ between them.
Grouped with the corresponding action labels $\mathfrak{a}_p$ and $\mathfrak{a}_c$ of action motions $\mathscr{S}_p$ and $\mathscr{S}_c$, we form the training input $[\mathscr{S}_p,\ \mathscr{T},\ \mathscr{S}_c]$, $\mathfrak{a}_p$ and $\mathfrak{a}_c$.

\noindent\textbf{Test data for MACVAE.}
MACVAE requires four test inputs: previous action motion, action label, and desired length $l_t$ and $l_c$ of generated motions.
We sample continuous motion and corresponding action labels $[\mathscr{S}_p,\ \mathscr{T},\ \mathscr{S}_c]$, $\mathfrak{a}_p$, and $\mathfrak{a}_c$.
$\mathscr{S}_p$ and $\mathfrak{a}_c$ are used as a previous action motion and action label, respectively, for the test time inputs.
Lengths of $\mathscr{T}$ and $\mathscr{S}_c$ are used for $l_t$ and $l_c$, respectively.

\noindent\textbf{Test data for MultiAct.}
MultiAct takes a series of action labels $\{a_k\}$ for timestep $k = 1, ..., n$, and lengths $l_{T_k}$, $l_{S_k}$ of generated motions $T_k$, $S_k$, respectively.
The first $a_1$ is randomly sampled in $A$.
Then, we inductively sample $a_{k+1}$ from the set $\{a_{k+1} \in A | \ (a_k, a_{k+1}) \ \text{in test set}\}$.
Lengths $l_{T_k}$ and $l_{S_k}$ are determined by $a_k$: mean length of action motions of label $a_k$, and mean length of transitions that comes before the action motion of $a_k$, respectively.

\subsection{Evaluation metrics}
As briefly introduced in manuscript, we use frechet inception distance \textbf{(FID)}, action recognition accuracy \textbf{(Acc.)}, diversity \textbf{(Div.)}, and multimodality \textbf{(Multimod.)} as the measurement of the quality of the generated motions, following previous works~\cite{petrovich21actor,chuan2020action2motion}.
We train an RNN action recognition model on action motions of the training set, which takes a sequence of 3D joint coordinates to calculate the action recognition accuracy directly or to use it as a feature extractor for other metrics.
We use the recognition model from Chuan \textit{et al.}~\cite{chuan2020action2motion} that is used for the same purpose.
We also follow Chuan \textit{et al.} to repeat 20 experiments and report confidence interval at 95\% confidence with $\pm$ symbols for a fair comparison.

\noindent\textbf{Frechet inception distance (FID).}
FID represents a difference between two sets, measured by a distance between fitted Gaussian distributions.
The Gaussian distribution is obtained by fitting Gaussian parameters to human motion features, extracted by the pre-trained action recognition model.
$\text{FID}_\text{train}$ is FID between 1,060 (20 per each action label) generated motions and motions in the training set.
Likewise, $\text{FID}_\text{test}$ is FID between 1,060 generated motions and motions in the testing set.
In addition, $\text{FID}_\text{test}$ of transition is a measured distance between generated transitions and GT transitions in the testing set.

\noindent\textbf{Accuracy (Acc.).}
The action recognition accuracy (Acc.) represents how likely generated motions are classified to their action label by the pre-trained action recognition model.
To this end, we run the pre-trained action recognition model on the generated motions and compare the result with the GT action label.
Since the BABEL dataset includes a large amount of duplicated action labels such as (``stand'', ``stand with arms down'', ``stand still''), and (``walk'', ``walk back'', ``walk backward'', ``walk backwards''), we use the confusion matrix to evaluate the recognition accuracy in every experiment.
Details of the confusion matrix will be provided together with the released code.

\noindent\textbf{Diversity (Div.) and Multimodality (Multimod.).}
The diversity and multimodality represent the variance of the generated motion across all actions and within each action, respectively.
We extract features using the pre-trained action recognition model from the set of generated motions from various actions and calculate the extracted features' variance.
The multimodality is calculated similarly.
Please refer to Action2Motion~\cite{chuan2020action2motion} for further details.

\bibliography{aaai23}

\end{document}